\documentclass[10pt,twocolumn,letterpaper]{article}
\usepackage[pagenumbers]{cvpr}
\usepackage{graphicx}
\usepackage{amsmath}
\usepackage{amssymb}
\usepackage{booktabs}
\usepackage{flushend}
\usepackage{xspace}
\usepackage{xcolor}
\usepackage{color}
\usepackage{colortbl}
\usepackage[pagebackref,breaklinks,colorlinks]{hyperref}
\usepackage[capitalize]{cleveref}
\crefname{section}{Sec.}{Secs.}
\Crefname{section}{Section}{Sections}
\Crefname{table}{Table}{Tables}
\crefname{table}{Tab.}{Tabs.}
\graphicspath{{imgs/}}

\definecolor{turquoise}{cmyk}{0.65,0,0.1,0.3}
\definecolor{purple}{rgb}{0.65,0,0.65}
\definecolor{dark_green}{rgb}{0, 0.5, 0}
\definecolor{orange}{rgb}{0.8, 0.6, 0.2}
\definecolor{dark_orange}{rgb}{0.7, 0.6, 0.3}
\definecolor{red}{rgb}{0.8, 0.2, 0.2}
\definecolor{darkred}{rgb}{0.6, 0.1, 0.05}
\definecolor{blueish}{rgb}{0.0, 0.3, .6}
\definecolor{light_gray}{rgb}{0.7, 0.7, .7}
\definecolor{pink}{rgb}{1, 0, 1}
\definecolor{cyan}{rgb}{0., 1, 1}

\definecolor{checkyes}{rgb}{0.7, 1.0, 0.7}
\definecolor{crossno}{rgb}{1.0, 0.7, 0.7}

\definecolor{tabbestcolor}{rgb}{0.785, 0.851, 0.969}
\def \best {\cellcolor{tabbestcolor!85}}
\def \sbest {\cellcolor{tabbestcolor!30}}

\definecolor{mydarkblue}{rgb}{0,0.08,0.55}
\hypersetup{
    colorlinks=true,
    linkcolor=mydarkblue,
    citecolor=mydarkblue,
    filecolor=mydarkblue,
    urlcolor=mydarkblue
}

\renewcommand{\paragraph}[1]{\vspace{.2em}\noindent\textbf{#1}.}

\newcommand{\repast}{RePAST\xspace}
\begin{document}
\title{\repast: Relative Pose Attention Scene Representation Transformer}
\author{
    Aleksandr Safin$^1$
    \qquad Daniel Duckworth$^2$
    \qquad Mehdi S. M. Sajjadi$^2$\\
    $^1$Skolkovo Institute of Science and Technology
    \qquad $^2$Google Research, Brain Team
}
\providecommand{\RePoST}[0]{\repast}
\providecommand{\Sin}{\text{sin}}
\providecommand{\Cos}{\text{cos}}
\providecommand{\Frequency}{\omega}
\providecommand{\NumImages}{N}
\providecommand{\Image}{\mathbf{I}}
\providecommand{\Camera}{\mathbf{C}}
\providecommand{\Reals}{\mathbb{R}}
\providecommand{\FeatureMap}{\mathbf{F}}
\providecommand{\CNN}{\text{CNN}}
\providecommand{\PatchToken}{\mathbf{x}}
\providecommand{\Height}{\text{H}}
\providecommand{\Width}{\text{W}}
\providecommand{\Channels}{\text{C}}
\providecommand{\Ray}{\mathbf{r}}
\providecommand{\Query}{\mathbf{Q}}
\providecommand{\Key}{\mathbf{K}}
\providecommand{\Value}{\mathbf{V}}
\providecommand{\ToCoord}{\Pi}
\providecommand{\ConcatenateTo}{\oplus}
\providecommand{\BigConcatenateTo}{\bigoplus}
\providecommand{\PositionalEncoding}{\gamma}
\providecommand{\QueryRay}{\mathbf{q}}
\providecommand{\QueryVector}{\mathbf{v}}
\providecommand{\Processed}[1]{\tilde{#1}}
\maketitle
\begin{abstract}
The Scene Representation Transformer (SRT) is a recent method to render novel views at interactive rates.
Since SRT uses camera poses with respect to an arbitrarily chosen reference camera, it is not invariant to the order of the input views.
As a result, SRT is not directly applicable to large-scale scenes where the reference frame would need to be changed regularly.
In this work, we propose Relative Pose Attention SRT (\repast):
Instead of fixing a reference frame at the input, we inject pairwise relative camera pose information directly into the attention mechanism of the Transformers.
This leads to a model that is by definition invariant to the choice of any global reference frame, while still retaining the full capabilities of the original method.
Empirical results show that adding this invariance to the model does not lead to a loss in quality.
We believe that this is a step towards applying fully latent transformer-based rendering methods to large-scale scenes.
\end{abstract}

\section{Introduction}
\label{sec:intro}

Implicit neural scene representations are currently a highly active field of study \cite{tewari2021advances}.
With the rise of neural radiance fields~\cite{Mildenhall20eccv_nerf}, extensions and adaptations that are trained across scenes in an amortized fashion have gained popularity~\cite{Yu21cvpr_pixelNeRF, Trevithick21iccv_GRF}.
Taking a different, more implicit approach, the Scene Representation Transformer (SRT)~\cite{srt} trains a single model on large datasets while replacing the inefficient volumetric rendering with a direct light field parametrization of the space.
During training, each data point is a different scene which consists of $N$ views of the same local spatial region.
An encoder transforms these views into the set-latent scene representation (SLSR), which is then used in the decoder transformer to render novel views.

\begin{figure}[t]
    \centering
    \includegraphics[width=0.9\linewidth,trim={1cm 10cm 16.5cm 1.6cm},clip]{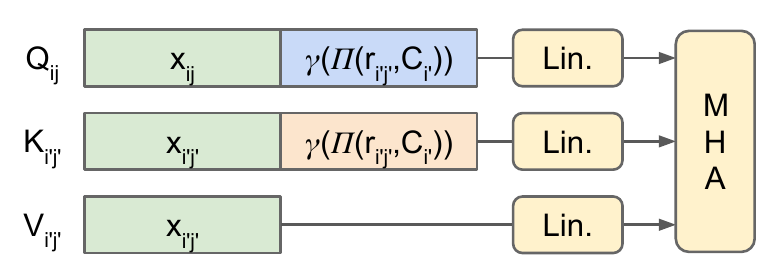}
    \caption{
    \textbf{\emph{Relative Pose Attention (RePA)}} --
    Both query and key are augmented with their pose relative to the camera $\Camera_{i'}$ belonging to the key token $\Key_i$.
    }
    \vspace{-2mm}
    \label{fig:repa}
\end{figure}

While SRT has achieved state-of-the-art novel view synthesis results in challenging settings, it has so far been restricted to single, local scenes with few views.
To this end, one major roadblock is its handling of camera poses:
Both input views and target views are parameterized by their camera origin and ray directions to inform the model of the position of each pixel from the input views that the model receives as input, or pixels of the target views that the model is queried at.
This leaves open the question of which reference frame to choose for the 3D pose parametrizations that are fed into the model.
In SRT~\cite{srt}, one of the input cameras is arbitrarily chosen to be the reference frame of the entire scene.
This particular choice leads to two problems:
First, the arbitrary choice of the reference camera leads to several possible parametrizations of the same exact scene, \ie the model has to learn this invariance during training.
Second, the reference frame has to be changed if SRT is to be applied to large scenes with many views, since the switch to the $K$ nearest input views would at times require the choice of the reference camera pose to be changed.
We find that in practice, the model does not learn to be fully invariant to the choice of the reference camera, leading to flickering artifacts when the reference camera is changed (see \cref{sec:exp:cycle}).
Making the model invariant to the choice of the reference camera is not trivial, since poses are central in the design of the SRT architecture: input view poses are concatenated at the very beginning to the RGB views (which are fed into the CNN), and target view poses are used as queries to the decoder to establish a correspondence to the input views that were used to produce the SLSR.

In this work, we present the \emph{Relative Pose Attention Scene Representation Transformer} (\repast), a method that incorporates reference-point invariant relative pose information between cameras and pixels inside the self-attention and cross-attention transformer layers.
In our experiments, we discover that \repast produces comparable or slightly better results then SRT in terms of PSNR, SSIM and LPIPS despite being invariant to the choice of a particular reference camera.

\begin{figure}[t]
    \centering
    \includegraphics[width=\linewidth,trim={0cm 9cm 10cm 0cm},clip]{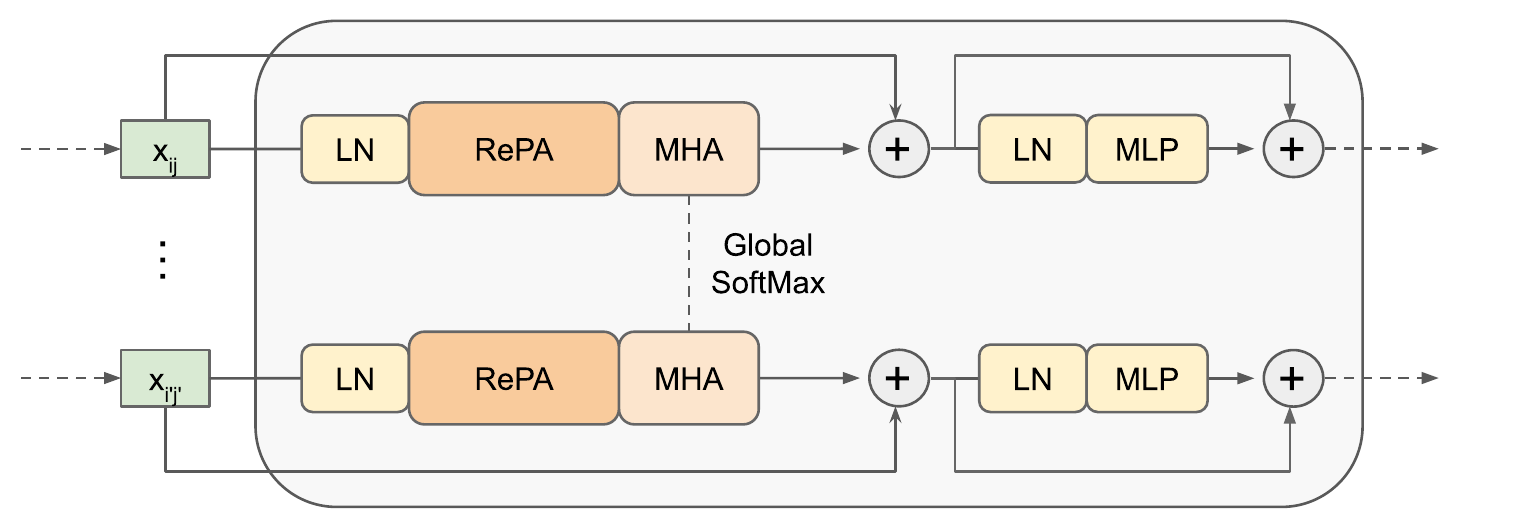}
    \caption{
    \textbf{\emph{\repast Encoder Layer}} --
    Each SLSR token self-attends into all other tokens using RePA.
    As in SRT's vanilla self-attention, softmax layers are computed globally over all tokens.
    }
    \label{fig:enc}
\end{figure}

\section{Method}
\label{sec:method}

We begin by describing SRT, the foundation upon which our method is built.
We demonstrate SRT's sensitivity to input view ordering and identify where in the model this sensitivity is introduced.
Finally, we introduce \RePoST, a coordinate-system invariant Scene Transformer model.

\subsection{Scene Representation Transformer}
Scene Representation Transformer~\cite{srt} (SRT) is a method for real-time novel view synthesis via set-latent scene representations.
At its core, SRT transforms a set of posed RGB images to a \emph{set-latent scene representation} (SLSR), a set of latent patch tokens in the spirit of ViT~\cite{vit}.
Among other things, this representation can then be used for tasks such as predicting the colors of 3D camera rays via a decoder module.
When using a light field decoder module, scenes can be encoded and novel views generated at real-time speeds.

\paragraph{Architecture}
SRT's architecture is divided into an encoder and a decoder.
The encoder is responsible for transforming posed RGB images into an SLSR.
This module takes the form of a CNN backbone, followed by a series of self-attention blocks.
To provide the model with a sense of 3D geometry, the origin and direction corresponding to each pixel's camera ray is provided alongside RGB color.
In practice, these values are further pre-processed by sinusoidal positional encodings.
Once the SLSR is generated, a light field decoder module is used to predict the color of a query camera ray.
As in the encoder, the query ray's origin and direction are featurized via sinusoidal positional encodings.
The decoder takes the form of a small number of cross-attention Transformer layers, followed by an MLP.
Given the small size of the decoder and lack of ray marching~\cite{Mildenhall20eccv_nerf}, novel views can be generated at over 100 frames per second on commodity hardware~\cite{srt}.

\paragraph{Coordinate Systems in SRT}
Unlike many recent methods in novel view synthesis~\cite{Tewari20egsr_neural_rendering}, SRT is only weakly biased towards respecting the rules of 3D geometry.
Rather than ray marching and volumetric rendering, SRT directly predicts color from a ray's origin and direction.
Beyond visual information provided by input images, the only mechanism by which the model derives 3D knowledge is in the representation of 3D quantities such as ray origin and direction.
Naturally, the choice of \emph{which} coordinate system has an impact on the model's predictions.
As demonstrated in \cref{sec:exp:cycle}, \emph{SRT is not invariant to choice of coordinate system}, since
a single, fixed canonical coordinate system is chosen before computation begins.
All camera rays, including those in the input images and in query renders, are represented in this system.
This impacts SRT's predictions both in the encoder, where camera rays corresponding to input images are provided; and in the decoder, where query rays are used.

\begin{figure}[t]
    \centering
    \includegraphics[width=\linewidth,trim={0cm 9cm 10cm 0cm},clip]{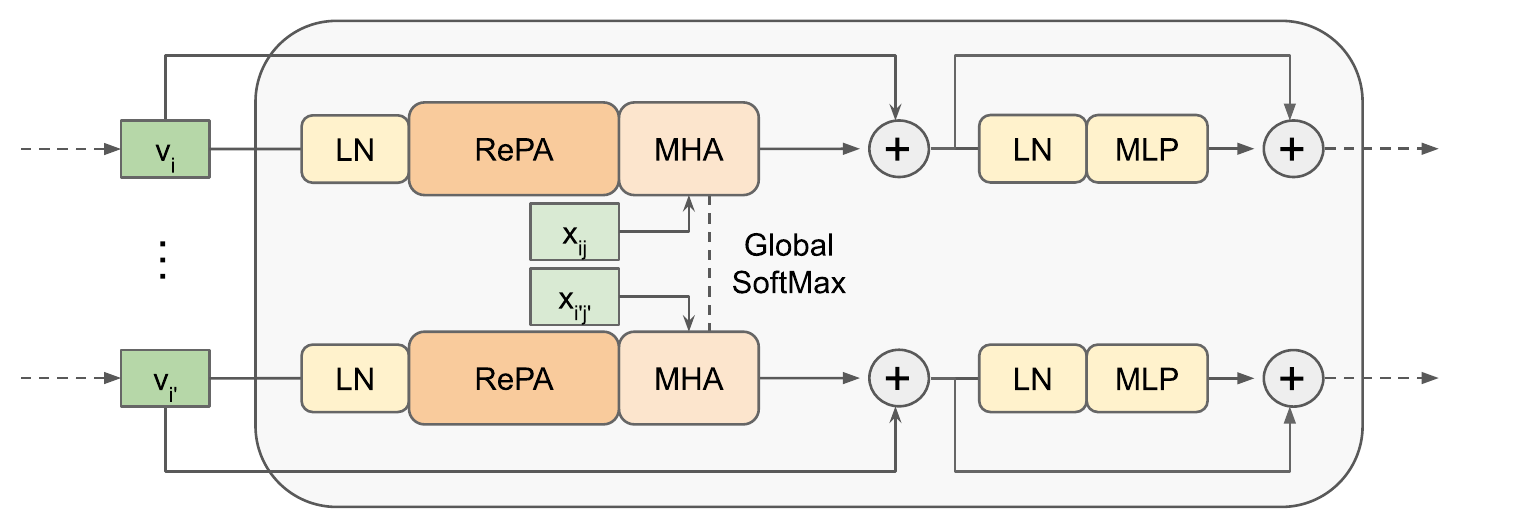}
    \caption{
    \textbf{\emph{\repast Decoder Layer}} --
    The decoder behaves similar to the encoder layers.
    Instead of self-attention, the target view queries cross-attend into the SLSR using RePA.
    The N decoding streams interact through the global softmax, and the results are averaged for a final MLP (not shown here) to produce the target RGB color.
    }
    \label{fig:dec}
\end{figure}

\subsection{\repast: Relative Pose Attention SRT}
We now present \RePoST, a coordinate system-invariant approach for SRT.
\RePoST introduces a series of targeted changes to SRT such that the desired invariance property is preserved.
The key idea behind \RePoST is the use of \emph{pairwise-relative coordinate systems}.
Rather than using a single, fixed coordinate system for all computations, the coordinate system varies depending on the quantities participating in pairwise interactions.
As \RePoST lacks a concept of a global reference frame, it is by definition invariant to changes in coordinate systems.

\paragraph{Encoder}
Similar to SRT, \RePoST adopts a CNN backbone followed by a series of Transformer blocks.
However, camera ray origins and directions are \emph{not} concatenated to RGB images.
Instead, the CNN operates on RGB alone and a patch's (average) origin and direction is introduced in \RePoST's encoder block.
The key difference between SRT's and \RePoST's encoder block is the use of \emph{Relative Pose Attention (RePA)}, see \cref{fig:repa,fig:enc}.
In SRT's self-attention layer, a set of latent patch embeddings is converted to key, query, and value vectors via linear layers.
As in Transformers, each query is compared against against each key, which in turn results in a weighted sum of value vectors via attention.
Critically, the \emph{same} query is compared against each key, resulting in the familiar expression,
\begin{align}
  \text{Attention}(Q, K, V) = \text{softmax} \left(
    \frac{Q K^T}{\sqrt{d_k}}
    V
  \right)
\end{align}
In RePA, each key and query is augmented with \emph{pose information} in the form of positionally-encoded patch origins and directions immediately before a comparison.
Unlike SRT, the coordinate system for each comparison varies.
Namely, the camera coordinate system relative to the key is employed.

Formally, consider a collection of posed input images $\{ \Image_i \in \Reals^{H \times W \times 3} \}$ with corresponding extrinsics $\{ \Camera_i \in \text{SE}(3) \}$ for a single scene.
For the sake of exposition, assume that all images share the same camera intrinsics.
The CNN backbone converts each image to a set of feature maps or, equivalently, a set of patch tokens,
\begin{align}
  \FeatureMap_i &= \CNN(\Image_i) \in \Reals^{\Height' \times \Width' \times \Channels}
    && i = 1 \ldots \NumImages
    \\
  \PatchToken_{ij} &= \FeatureMap_i \left[ j \text{ // } \Width', j \text{ \% } \Width' \right]
    && j = 1 \ldots \Height' \Width'
\end{align}
We associate each latent patch token $\PatchToken_{ij}$ with a corresponding camera ray $\Ray_{ij}$, which starts at camera $\Camera_i$'s origin and passes through the center of the corresponding patch $j$.

The input to a \RePoST encoder block is the set of patch tokens $\{ \PatchToken_{ij} \}$ and corresponding camera rays $\{ \Ray_{ij} \}$.
Consider a single self-attention comparions between positions query $ij$ and key $i'j'$.
We construct query embedding $\Query_{ij}$ by realizing $\Ray_{ij}$'s origin and direction in the coordinate system of camera $i'$,
\begin{align}
  \Query_{ij} =
    \PatchToken_{ij}
    \ConcatenateTo
    \PositionalEncoding \left(
      \ToCoord \left( \Ray_{ij}, \Camera_{i'} \right)
    \right)
\end{align}
where $\ConcatenateTo$ represents vector concatenation, $\PositionalEncoding$ represents sinusoidal positional encoding, and $\ToCoord(\Ray, \Camera)$ constructs the origin and direction of a ray $\Ray$ in the local coordinate system of camera $\Camera$.
Similarly, the key embedding $\Key_{i'j'}$ is defined as,
\begin{align}
  \Key_{i'j'} =
    \PatchToken_{i'j'}
    \ConcatenateTo
    \PositionalEncoding \left(
      \ToCoord \left( \Ray_{i'j'}, \Camera_{i'} \right)
    \right)
\end{align}
Note that $\Ray_{ij}$ and $\Ray_{i',j'}$ are both expressed in the coordinate system of $\Camera_{i'}$.
Once the query-key compatibility matrix is constructed, \RePoST's encoder block proceeds as in SRT.
The result is a set of new patch tokens, which are then fed to the next encoder block.

\paragraph{Decoder}
Unlike the encoder, the decoder operates as a form of \emph{cross} attention between a query ray and the SLSR.
Whereas each encoder block updates the \emph{scene} representation, the decoder blocks update the \emph{query} representation.
The \RePoST decoder block diverges from the SRT decoder in two ways.
First, each query ray is internally represented not once, but $\NumImages$ times, once per input image.
Second, 3D rays are represented within pairwise-relative coordinate systems similar to the \RePoST encoder.
The \RePoST decoder block is shown in \cref{fig:dec}.
Consider a single query ray $\QueryRay$ and an SLSR $\{ \PatchToken_{ij} \}$; the following is trivially parallelizable across multiple queries.
In the first decoder block, $\NumImages$ local query vectors $\QueryVector_i$ are initialized, one per input image.
Similar to the \RePoST encoder, a query-key compatibility matrix is constructed via RePA, using $\QueryVector_i$ for all comparisons with respect to $K_{ij}, j = 1 \ldots \Height' \Width'$.
Multi-headed cross-attention~\cite{transformer} is then applied to obtain a single latent vector, $\Processed{\QueryVector}$.
Each local query vector is then added to this quantity via residual connections to construct $\Processed{\QueryVector}_i$.
The remainder of the block is identical to SRT.
The output is a set of local query vectors to be used in the next iteration.
After the final decoder block, the local query vectors are average-pooled and an MLP is applied, predicting the ultimate RGB color.

Crucially, neither encoder, nor decoder make any reference to an arbitrary global coordinate system.
As a result, the resulting SLSR is independent of coordinate system changes (see \cref{sec:exp:cycle}).

\begin{table}[t]
\centering
\small
\begin{tabular}{lcccccc}
	\toprule
	& $\uparrow$\,PSNR
	& $\uparrow$\,SSIM
	& $\downarrow$\,LPIPS \\
    \cmidrule(lr){2-4}
	{SRT~\cite{srt}} & 24.61 & 0.784 & 0.223 \\
	{\repast}
	& \best 24.89 & \best 0.794 & \best 0.202 \\
	{\repast-B}
	& \sbest 24.71 & \sbest 0.788 & \sbest 0.211 \\
	\bottomrule
\end{tabular}
\caption{
\textbf{Quantitative results} --
\repast modestly improves over SRT across all metrics.
Removing the relative camera injection in the Decoder (\repast-B) leads to slightly lower quality, see \cref{sec:exp:quantqual}.
}
\label{tab:quantitative}
\end{table}

\section{Experimental Results}
\label{sec:exp}

\repast is built on the improved SRT variant presented by~\cite{osrt}.
Unless stated otherwise, all hyper parameters are as reported in that work.
As a training dataset, we followed prior work by using the MultiShapeNet-Hard dataset (MSN)~\cite{osrt}.
It consists of 1\,M scenes for training and we used the first 128 test scenes for evaluation.
Each test scene contains novel objects from ShapeNet~\cite{chang2015shapenet} that have not been observed at training time.
Further, the object arrangement is randomized for each scene.
The image resolution of this rendered dataset is 128$\times$128 and all models were trained for 4\,M steps~\cite{srt}.

\subsection{Reference Frame Invariance}
\label{sec:exp:cycle}

Contrary to SRT, \repast is by design invariant to the choice of the reference camera, since it uses only pair-wise relative information regarding camera positions.
We investigate the extend to which SRT is sensitive to changes in the choice of the reference camera by rendering the same target view from the same set of 5 input views, but cycle through the latter to produce an animated gif.
The resulting video for SRT is available at \href{https://ibb.co/BGQ9TSk}{ibb.co/BGQ9TSk}.
It is evident that SRT has learned to be mostly invariant to the choice of the reference camera, but still produces significant changes in local geometry when the reference camera is changed.
\repast does not have a choice for a reference camera, so its results are nearly invariant to the order of the input views: \href{https://ibb.co/gW9cGp2}{ibb.co/gW9cGp2}.

\subsection{Quantative and qualitative evaluation}
\label{sec:exp:quantqual}

We compare \repast with SRT on for the novel view synthesis task.
\cref{tab:quantitative} shows quantitative results.
The results show that \repast does not lose quality compared to the baseline, but achieves slightly better numbers.
Qualitative results are shown in \cref{fig:qualitative}.
Visually, differences between the models are subtle, though \repast produces slightly sharper images than the SRT baseline.

\newcommand\mspic[1]{
    \includegraphics[width=0.9\linewidth]{imgs/qualitative/#1}
}

\begin{figure}[t]
    \centering 
    \setlength{\tabcolsep}{0.5mm}
    \def\arraystretch{0.4}
    \begin{tabular}{cl}
    \rotatebox{90}{\hspace{4mm}\small Input} & \mspic{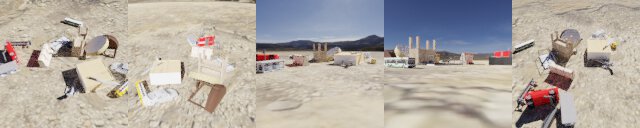} \\
    \rotatebox{90}{\hspace{3mm}\small Target} & \mspic{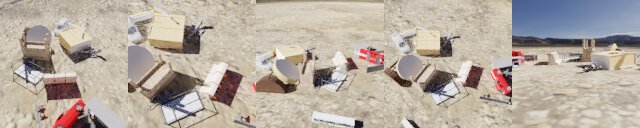} \\
    \rotatebox{90}{\hspace{4mm}\small SRT} &\mspic{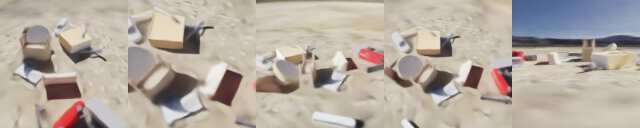} \\
    \rotatebox{90}{\hspace{2mm}\small\repast} &\mspic{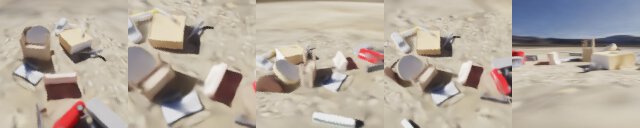} \\
    \rotatebox{90}{\hspace{0.4mm}\small\repast-B} & \mspic{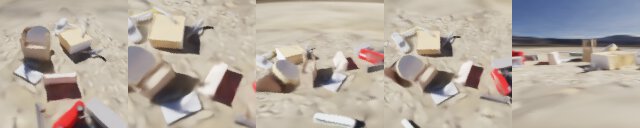} \\
    \end{tabular}
    \caption{
    \textbf{Qualitative results on MSN} --
    \repast produces slightly higher-quality results compared to SRT.
    }
\label{fig:qualitative}
\end{figure}

\paragraph{Ablation}
In the \repast Decoder model, queries already contain the full (relative) pose information.
Because of this, it is not strictly necessary to inject the pairwise relative pose information into the decoder attention mechanism anymore.
We also study this variant, dubbed \repast-B in \cref{tab:quantitative} and \cref{fig:qualitative} and find that this change leads to a small drop in performance compared to \repast, showing that the inclusion of pairwise relative information in the Decoder attention is, while not strictly necessary, still helpful to the model.

\section{Related Work}
\label{sec:relatedworks}

\repast builds upon SRT~\cite{srt}, which requires the use of a scene-global coordinate frame.
Presented in the same work, UpSRT does not require \emph{input} camera pose information at test time, but to render novel views, UpSRT still requires target pose information \emph{relative} to an arbitrarily chosen input view.
More recently, RUST~\cite{sajjadi2022rust} drops camera pose information altogether by learning unstructured latent representations for the camera poses.

Most methods based on NeRF~\cite{Mildenhall20eccv_nerf} use absolute pose, since a single model is directly trained to overfit to a single scene.
Methods that generalize NeRF across scenes regularly make arbitrary choices for the reference camera, for example PixelNeRF~\cite{Yu21cvpr_pixelNeRF}, or MVSNeRF~\cite{Chen21iccv_MVSNeRF}, which uses a 3D CNN on a voxel grid that is aligned with normalized device coordinates.
NeRF-VAE~\cite{Kosiorek21icml_NeRF_VAE} uses a fully latent scene representation, but makes use of an absolute scene coordinate system, which does not easily generalize to more complex, unbounded scenes.
Some popular fully latent 3D models trained through novel view synthesis only allow for one input camera while using relative cameras \cite{gfvs}.
It should be noted that when using only a single input camera, SRT is similar to \repast in the sense that all target poses are relative to the single input view.

Introducing camera reference frame invariance to transformers for novel view synthesis is, to the best of our knowledge, new.
There has been recent works on introducing 2D position invariance to Slot Attention~\cite{slotattention} models~\cite{biza2023invariant} by directly using the attention patterns.
Vector Neurons~\cite{deng2021vector} proposed replacing common neural network building blocks such as MLPs and ReLU units by translation and rotation equivariant or invariant counterparts, but does not include proposals for pose-invariant attention or transformer layers.

\section{Conclusion and Future Work}
\label{sec:conclusion}

We propose a natural extension to SRT that allows it to be invariant to the order of the input cameras or the arbitrary choice of a particular reference frame.
We inform the model of the pairwise relative camera poses inside the Transformer's attention mechanism in a natural way.
Experiments show that \repast does not lead to worse results despite adding invariance to the model.

We believe that this work is a stepping stone towards applying SRT to large-scale scenes where the number of images is too large to be processed in a single feed-forward pass.
In such settings, only the closest few input views can be processed at a time (\eg, \cite{Yu21cvpr_pixelNeRF}), making it necessary to replace the farthest input view with the now closest new input view.
We have shown that \repast handles this use case much more gracefully compared to SRT.
Further, since the proposed technique for injecting pairwise relative camera pose information the attention mechanism is fairly general, it could in principle be applied to further Transformer-based methods (other than SRT) in future work.

\section*{Acknowledgements}
We thank Aravindh Mahendran, Etienne Pot, Klaus Greff, and Ricardo Martin-Brualla for their advice.
{\small
\bibliographystyle{ieee_fullname}
\bibliography{main}
}
\end{document}